\pdfoutput=1

\documentclass[11pt]{article}

\usepackage{acl}

\usepackage{times}
\usepackage{latexsym}
\usepackage{graphicx}
\usepackage{booktabs}
\usepackage{subcaption}
\usepackage{amsmath}
\usepackage{amssymb}
\usepackage{multirow}

\usepackage[T1]{fontenc}

\usepackage[utf8]{inputenc}

\usepackage{microtype}

\title{CI w/o TN: Context Injection without Task Name for Procedure Planning}

\author{Xinjie Li \\
  {\tt xql5497@psu.edu} }

\begin{document}
\maketitle





\begin{abstract}

This paper explores the challenge of procedure planning in instructional videos, which involves creating goal-directed plans based on visual start and goal observations from videos. Previous research has tackled this problem with gradually weaker training supervision, from heavy intermediate visual observations or language instructions to task class supervision. However, with the advent of large language models, even given only the task name, these models can produce a detailed plan. In this study, we propose a much weaker setting without task name as supervision, which is not currently solvable by existing large language models since they require good prompts with sufficient information. Specifically, we hypothesize that previous intermediate supervisions can serve as context information, and we use captions of visual start and goal observations as a much cheaper form of supervision. This approach greatly reduces the labeling cost since the captions can be easily obtained by large pre-trained vision-language foundation models. Technically, we apply BLIP to generate captions as supervision to train the context feature with contrastive learning loss. Afterward, the context feature is fed into the generator to aid in plan generation. Our experiments on two datasets with varying scales demonstrate that our model can achieve comparable performance on multiple metrics, which validates our hypothesis.

\end{abstract}

\section{Introduction}



\begin{figure}[t]
    \centering

    \begin{subfigure}{1.1\columnwidth}
      \centering   
      \includegraphics[width=1\columnwidth]{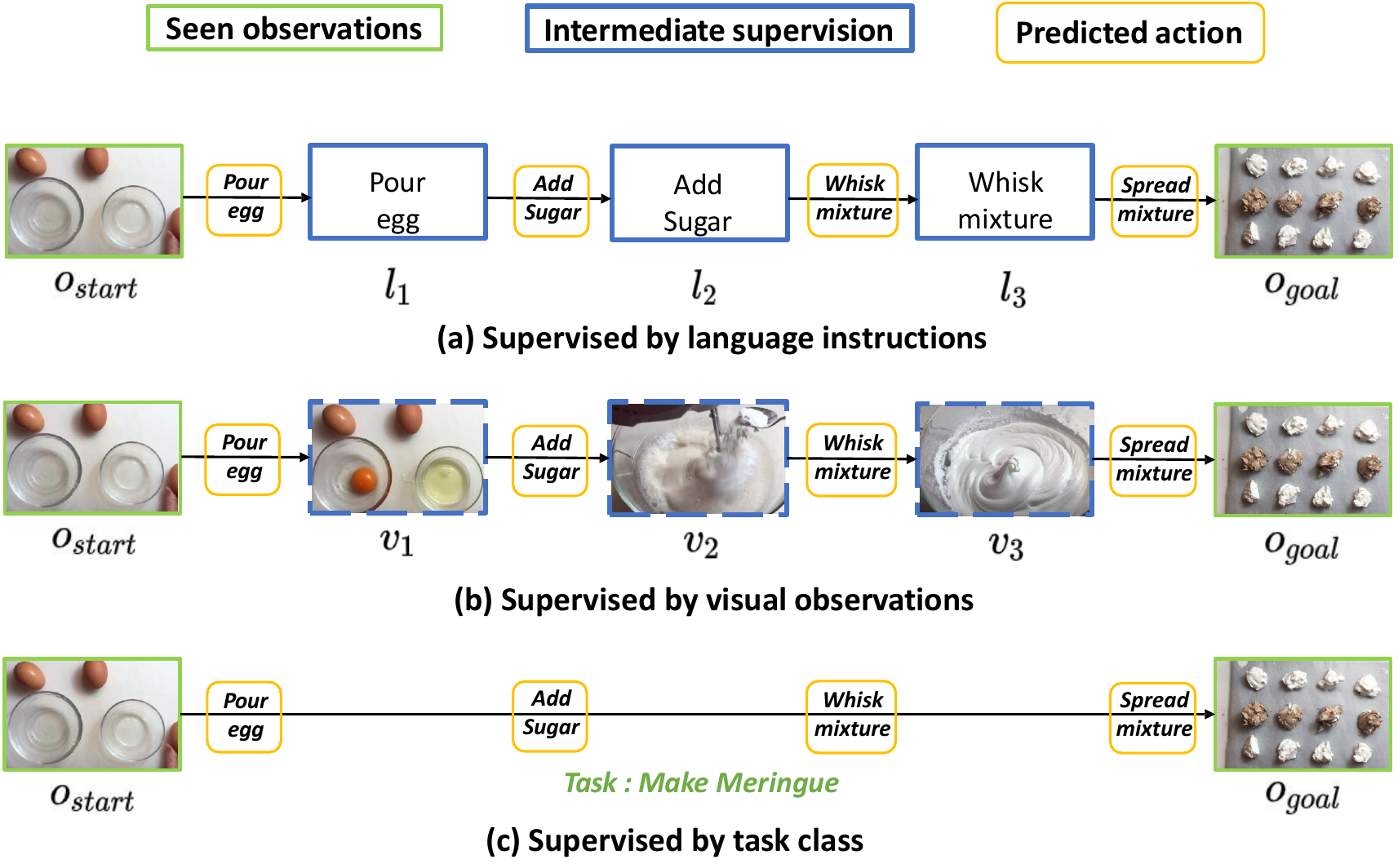}
    \end{subfigure}

    \begin{subfigure}{1.1\columnwidth}
      \centering   
      \includegraphics[width=1\columnwidth]{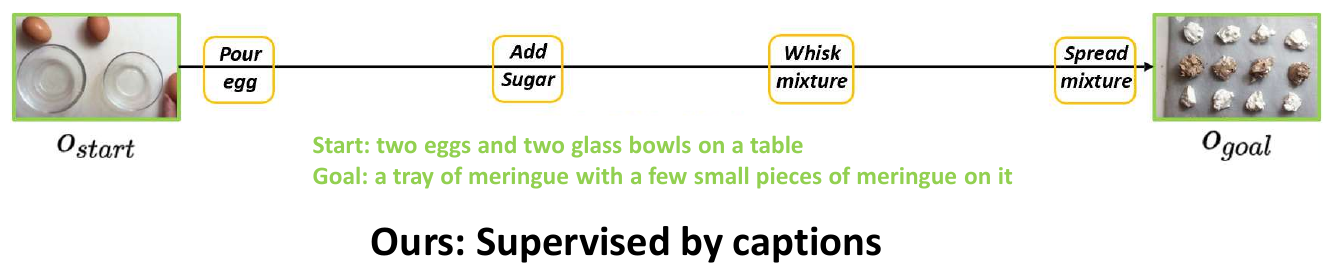}
    \end{subfigure}

\caption{
Illustration of Different Types of Supervision in Procedural Learning: Language Instructions, Intermediate Visual Observations, and Task Class. In our project, we have removed all the intermediate supervision and instead rely on captions of visual start and goal as context information. This figure is modified from the work of ~\citet{wang2023pdpp}.
}
\label{fig:previous}
\end{figure}


In the current era of large-scale foundation models, there has been increased attention toward addressing hardcore problems that remain unsolved by these large models. One such problem is long-term video understanding in computer vision, which requires sophisticated long-range temporal reasoning that foundation models struggle with. Instructional video analysis, with its natural semantic procedure relationships across the video, has emerged as a promising area for addressing long-term video understanding. 


In particular, procedure planning, which aims to predict a sequence of intermediate action steps as a plan to transform a start visual observation into a goal state, has been a challenging task in instructional video analysis. Initially, this task is proposed to be trained with intermediate visual observations and action steps (Fig.~\ref{fig:previous} (b) )~\cite{procedure2020}, but later weakly-supervised paradigm using language instructions as supervision is introduced (Fig.~\ref{fig:previous} (a) ) ~\cite{zhao2022p3iv}. More recently, a proposal to train the model with only the task name and action steps is put forth (Fig.~\ref{fig:previous} (c) ) \cite{wang2023pdpp}.

\begin{figure}[t]
    \centering
\includegraphics[width=1\columnwidth]{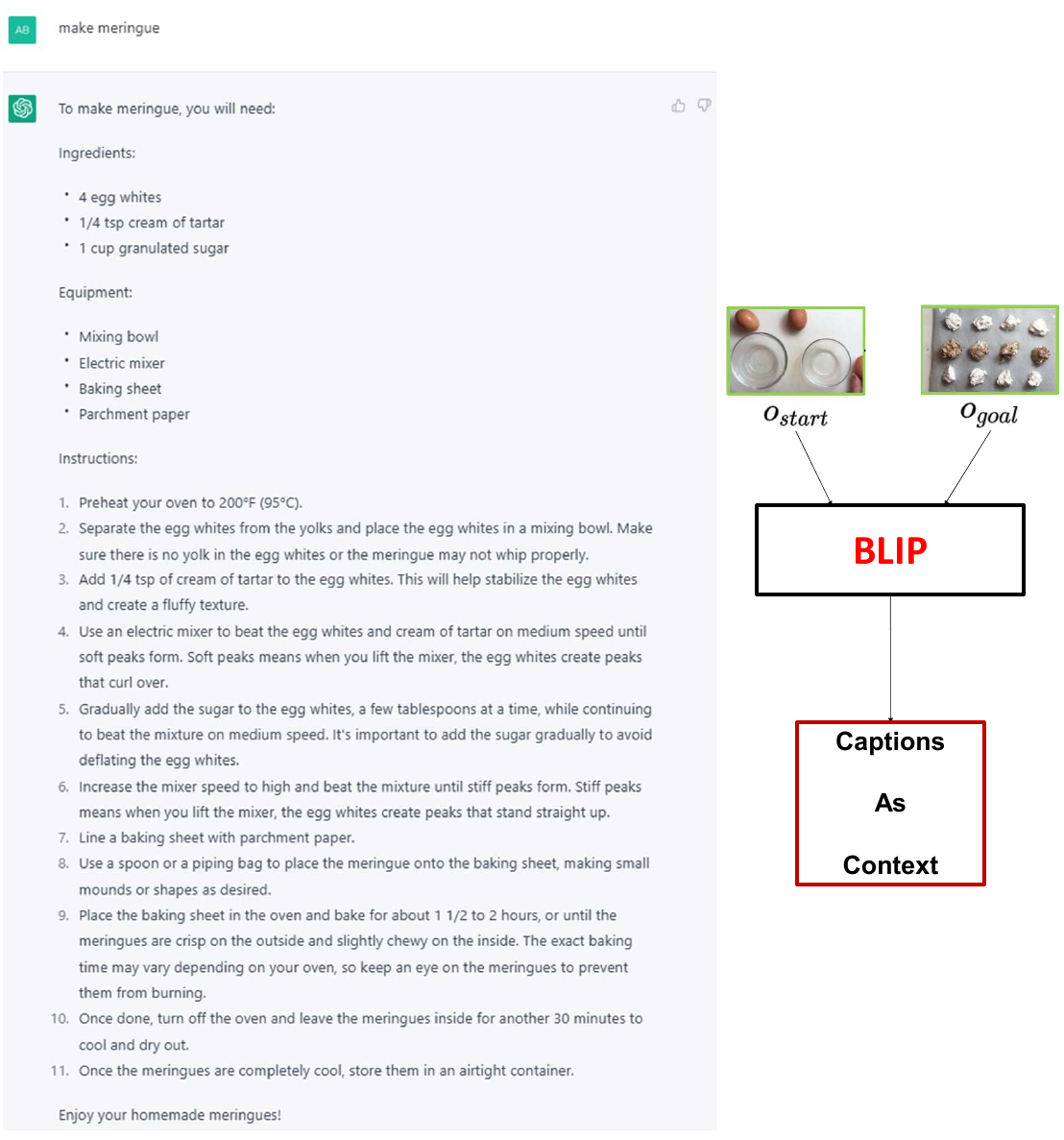}
\caption{
Motivations for our approach. On the left, we observe that when provided with a task name, ChatGPT can generate a detailed plan. However, a proper prompt containing sufficient information is always required for ChatGPT to retrieve knowledge from large language models. In our project, the model is not aware of the task name, which makes the task more challenging and currently unsolved by large language models. On the right, we hypothesize that all the intermediate supervisions, such as language instructions, visual observations, and task class, serve as context information for plan generation. Based on this assumption, we propose a cost-effective approach to obtain context information by generating captions for visual start and goal observations using a pre-trained BLIP model~\cite{li2022blip}.
}
\label{fig:motivation}
\end{figure}

However, in the context of large language models, it has been observed that these models can generate more detailed plausible plans even when provided with only the task name as supervision, as shown in Fig.~\ref{fig:motivation}. This intriguing phenomenon has motivated us to explore a much weaker setting without a task name as supervision, since a proper prompt, such as a task name, is always needed to retrieve knowledge from large language models.


To this end, we believe that reasonably weaker supervision than intermediate visual observations or language instructions, and task class is needed for our new setting. Through careful analysis of the role of previous intermediate supervisions in procedure learning, we hypothesize that these supervisions provide the model with context information. This has led us to consider alternative, cheaper sources of context supervision.


Inspired by recent work in video analysis, specifically in the use of captions~\cite{wu2022cap4video}, we propose to use captions of the visual start and goal observations as the training supervision. There are two main reasons for this choice. First, captions can be easily obtained by zero-shot image captioning with knowledge from web-scale pre-trained models, such as BLIP~\cite{li2022blip}, as shown in Fig.~\ref{fig:motivation}. Second, in the context of the procedure learning task, the start and goal observations contain the most significant context information, including ingredients, equipment, and the final product. Furthermore, to avoid additional inference costs, we only use the captions as supervision to train the context feature and inject it into the generator. In this way, the inference process remains unchanged from previous methods, but with the inclusion of better context information.

In summary, our contributions are as follows:
\begin{itemize}
    \item We propose a novel approach for exploring weaker supervision in the task of procedure planning in instructional video analysis which has not been addressed by large language models. 
    
    \item By leveraging captions of the visual start and goal observations as supervision, we aim to provide context information to the model for generating detailed plausible plans.
    
    \item Extensive experiments and evaluations are conducted to validate the effectiveness of our approach.
\end{itemize}

\section{Related Work}



\subsection{Procedure Planning in Instructional Videos}
Goal-conditioned planning has traditionally been focused on physical environments such as robotic motion planning \cite{kaelbling1993hierarchical, florensa2018automatic, ghosh2018learning} and human pedestrian trajectory planning \cite{mangalam2020not}. Recently, there has been a growing interest in the task of procedure planning from instructional videos \cite{procedure2020}. Various methods have been proposed for this task, including the use of recurrent neural networks (RNNs) \cite{procedure2020}, transformers \cite{sun2021plate}, and adversarial policy planning \cite{bi2021procedure}. These methods have all utilized two branches and strong visual supervision. In contrast,~\citet{zhao2022p3iv} adopt a non-autoregressive transformer-based architecture that directly models the actions. Moreover, they use low-cost weak supervision in the form of natural language instructions, as opposed to the "expensive" visual observations used in all existing approaches. Recently, ~\citet{wang2023pdpp} frame this problem as a distribution fitting task, where they model the distribution of the entire intermediate action sequences using a diffusion model. By doing so, they transform the procedure planning problem into a sampling process from this learned distribution. Furthermore, they eliminate the need for costly intermediate supervision, such as visual observations or language instructions, and instead leverage task labels from instructional videos as a more economical form of supervision. 

However, in the realm of large language models, these models possess the ability to generate highly detailed and plausible plans, even when they are only provided with minimal supervision in the form of task names. To this end, we propose a more challenging setting with captions of the visual start and goal observations as supervision.

\subsection{Zero-shot Image Captioning}
Within the realm of natural language processing, transformer-based Generative Pre-trained Transformer (GPT) models~\cite{GPT,GPT3} have emerged as a successful approach for generating text from prompts, leveraging extensive training on large-scale text corpora. Similarly, CLIP~\cite{clip}, a vision-language alignment model that has been trained on a massive dataset of 400 million image-text pairs, has exhibited remarkable zero-shot performance on vision tasks.

In recent works, ZeroCap~\cite{tewel2022zerocap} has proposed an innovative method that leverages both CLIP and the GPT-2 language model to generate textual descriptions of input images, showcasing the ability to utilize knowledge from both models in a truly zero-shot manner without the need for re-training or fine-tuning of model parameters. ~\citet{wu2022cap4video} further extend the video extension of ZeroCap, harnessing the power of CLIP and GPT-2 to generate informative and coherent captions for videos in a zero-shot manner, without the need for any further training.

Inspired by the above methods, we apply captions of the visual start and goal
observations as context information. Specifically, the captions are obtained by the BLIP model~\cite{li2022blip} which has been shown to achieve impressive zero-shot performance on image captioning tasks.



\begin{table*}[t]
\centering
\begin{tabular}{@{}llll@{}}
\toprule
            & \#Videos  & \#Procedures & \#Actions/Video  \\ \midrule
   CrossTask \cite{CrossTask}      & 2750 & 18 & 7.6 \\
COIN~\cite{COIN}  & 11827 & 778 & 3.6 \\ \bottomrule
\end{tabular}
\caption{Data statistics.}
\label{tab:dataset}
\end{table*}

\section{Data}


To assess our model's performance, we use three distinct instructional video datasets: CrossTask \cite{CrossTask}
and COIN~\cite{COIN}.
CrossTask features 2750 videos, with 18 different procedures and an average of 7.6 actions per video. 
COIN is the largest dataset in our evaluation, containing 11827 videos, 778 procedures, and 3.6 actions per video. The procedures depicted in these datasets vary widely, including tasks such as \textit{Make Taco Salad} and \textit{Change Car Tire}.

We adopt the 70\%/30\% split used in previous work \cite{procedure2020} to create our train/test sets, with 20\% of the training data reserved for validation. We also follow the data pre-processing steps outlined in the original procedure planning paper \cite{procedure2020} to select $\{start, goal\}$ observations and curate the dataset into plans covering different time horizons.

Data statistics are shown in Table~\ref{tab:dataset}.

\section{Model}

\begin{figure*}[t]
    \centering
    \includegraphics[width=1.\textwidth]{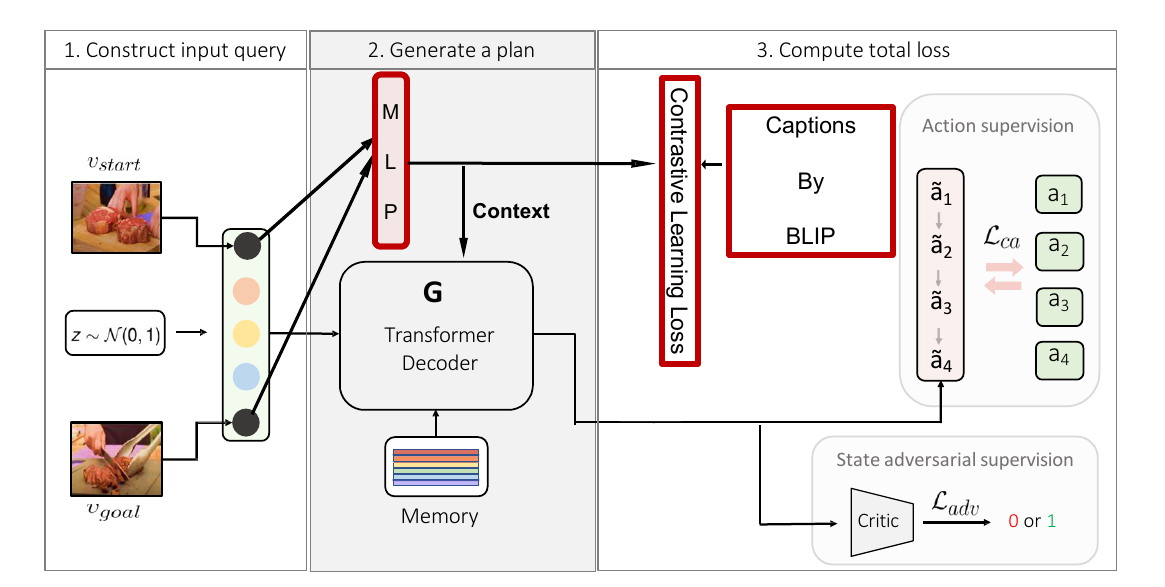}
    \caption{Overview of our model. Initially, visual observations of the starting and target states (represented by black nodes) are incorporated into the sequence of learned queries (represented by colored nodes). Then stochastic noise is injected into the resulting input sequence. Next, this input is fed into the transformer decoder, which interacts with the global memory to create executable procedural plans. Along with the memory, the context information transformed from the visual start and goal is injected into the transformer decoder. Subsequently, action vectors are generated, and multiple loss functions are employed, to train the model. Specifically, the context information is trained with contrastive learning loss and supervised by captions generated via BLIP~\cite{li2022blip}. Note that this figure is modified from~\cite{zhao2022p3iv} and our modifications are highlighted with red boxes.}
    \label{fig:model}
\end{figure*}

\begin{figure}[t]
    \centering
\includegraphics[width=1.05\columnwidth]{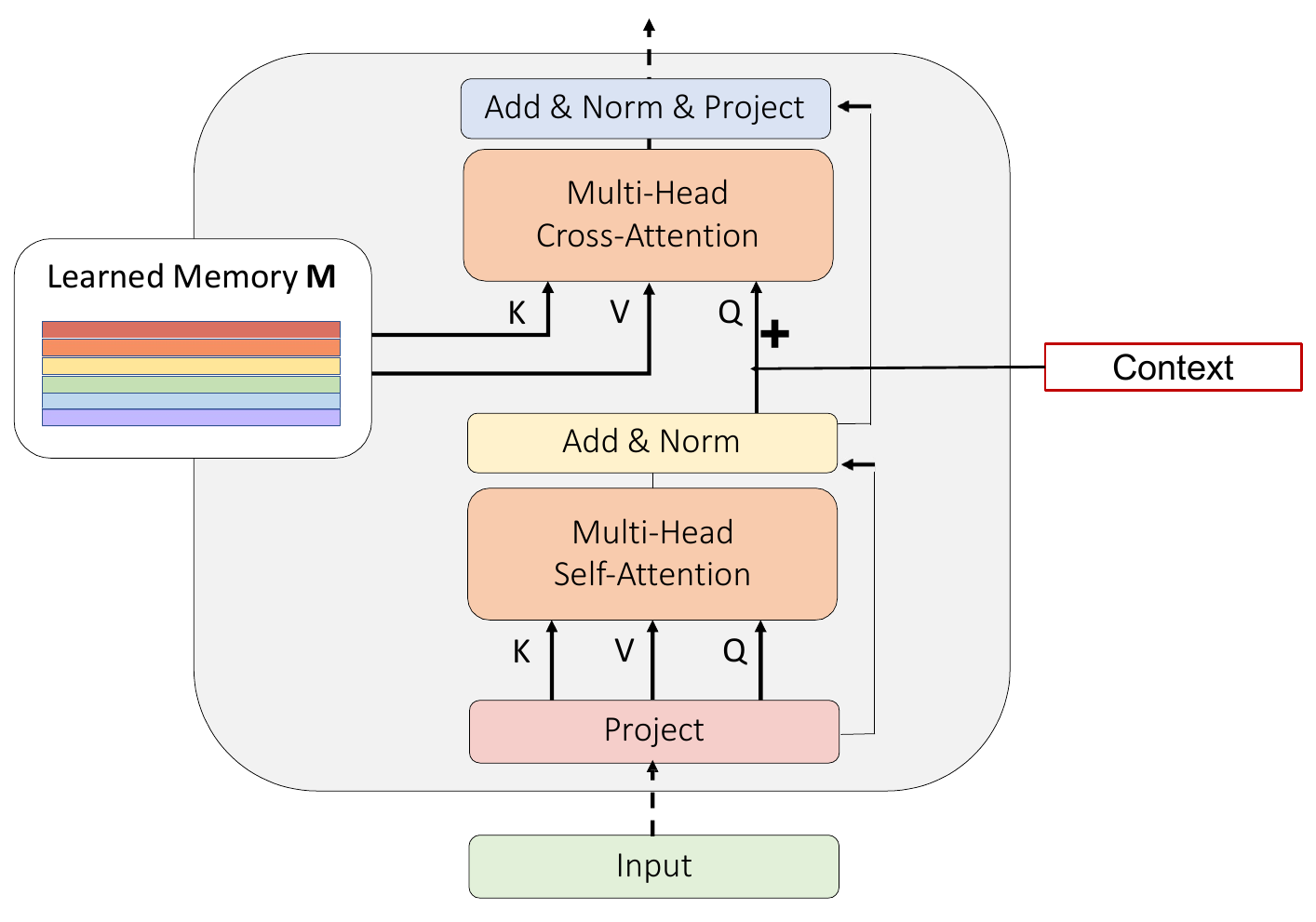}
    \caption{Context augmented transformer decoder. In this architecture, the learned memory bank provides the multi-head cross-attention with K and V. In addition, the learned context feature is concatenated with the input as Q. Note that this figure is modified from~\cite{zhao2022p3iv} and our modification is highlighted with the red box.}
\label{fig:attention}
\vspace{-0.7cm}
\end{figure}

\subsection{Overview}
Given an initial state denoted as $v_{start}$ and a target state represented as $v_{goal}$, our task of procedure planning aims to generate a plan comprising $T$ action steps, denoted by $\tilde{\pi} = \tilde{a}_{1:T}$, that facilitate the transition from the starting state $v_{start}$ to the target state $v_{goal}$.

As depicted in Fig.~\ref{fig:model}, in addition to $v_{start}$ and $v_{goal}$, a random noise vector, $\mathit{z} \sim \mathcal{N}(0, 1)$, is also fed into the generator $G$ as a stochastic component in the context of generative adversarial learning. Subsequently, an MLP layer is employed to generate the context feature by concatenating $v_{start}$ and $v_{goal}$, which is supervised by captions. The context feature, along with the memory feature, is then input into the generator $G$ to generate realistic action sequences. 

Adversarial training is employed in our approach, where the generator $G$ is trained to generate sequences while the critic $C$, modeled by a simple MLP, provides the supervisory signal. Specifically, we pass the output of our transformer, $G$, concatenated along the temporal dimension, to the critic $C$, which produces a value ranging from $0$ to $1$, indicating its ability to discriminate between the predicted and ground truth sequences.

\subsection{Context-augmented Generator}

In our work, we utilize a non-autoregressive transformer decoder architecture to implement the generator $G$. Apart from the learnable queries that are augmented with the start and goal observations, our transformer decoder takes two additional inputs, namely the context feature and the learned memory feature, to generate action predictions.

\subsubsection{Context Input} 
To minimize computational overhead during the inference stage, we employ a simple MLP layer to generate the context feature. As illustrated in Fig.\ref{fig:model}, we adopt a similar design as CLIP~\cite{clip}, which constructs a shared feature space through contrastive learning between the captions and the context feature. This allows for substantial augmentation of the context feature with the captions. Subsequently, the context feature is injected into the cross-attention module, as shown in Fig.\ref{fig:attention}. Drawing inspiration from recent advancements in visual attention\cite{shi2023top}, we simply concatenate the context feature to the query tokens.

\subsubsection{Memory Input} 
In order to further boost performance and ensure stable training, we introduce a learnable memory bank in our approach. This memory bank is read-only and shared across all examples in the dataset, being applied in all cross-attention layers. As illustrated in Fig.~\ref{fig:attention}, the learned memory serves as both key and value, enabling the model to access global historical information. This additional memory mechanism enhances the model's ability to capture long-range dependencies and contextual information, leading to improved performance and robustness in our approach.

\subsubsection{Context-augmented transformer decoder} 
Our proposed architecture is composed of a stack of standard transformer decoder blocks, each of which has access to the learnable memory and context feature. In particular, the context-augmented transformer block comprises two key operations. Firstly, the input is processed through the self-attention operation. Secondly, the cross-attention module incorporates input queries and context information to attend to the learnable memory and generate the output. Notably, the input to self-attention in the first transformer block corresponds to the query.

It is important to highlight that all cross-attention blocks have access to the same memory and context, creating a shared knowledge base. Intuitively, the memory module can be viewed as a collection of learnable plan embeddings that are shared across the entire dataset, allowing for consistent and coherent action generation.

To further elaborate, our transformer decoder is constructed as a stack of multiple context-augmented blocks, each contributing to the overall context-awareness and information integration. Additionally, we append an output MLP head after the final decoding layer to generate the predicted action steps, enabling the model to produce accurate and coherent action sequences for task planning.

\subsection{Training and Inference}
\subsubsection{Training}
During the training process, our approach involves predicting the plan $\tilde{\pi}$ given the start state $v_{start}$ and the goal state $v_{goal}$, and optimizing the entire network using three loss functions.

To train the context feature, we employ contrastive learning to establish a shared feature space with captions. Specifically, for each feature predicted by the MLP head, we use the corresponding ground truth caption embedding as the positive example, and all other embeddings in the caption database as negative examples. The contrastive loss is formulated as follows:

\begin{equation}
    \scalebox{1.15}{$\mathcal{L}_{c} = -\; \sum_{t=1}^{T} \left[ \text{log} \; \frac{\text{exp}(cap_{t} \cdot {cxt}_{t} )}{\sum_{j} \text{exp}( cap_{j} \cdot {cxt}_{t})} \right] $},
\end{equation}
where $cap_{t}$ denotes the ground truth caption embedding and ${cxt}_{t}$ represents the context feature.

In addition, we enforce the action prediction head, to generate sequences of action probabilities $\tilde{a}_t$ that correspond to the ground truth one-hot labels $a_t$. For this purpose, we employ the cross-entropy loss, defined as follows:

\begin{equation}
\scalebox{1.15}{$\mathcal{L}_{ca} = -\sum_{t=1}^{T} \, a_t \, \text{log} \, {\tilde{a}_t},$}
\end{equation}

As previously mentioned, we also incorporate adversarial training in our framework. We optimize the generator $G$ and the critic $C$ using an adversarial loss, given by:

\begin{equation}
\scalebox{1.15}{$\mathcal{L}_{adv} = \operatorname*{min}_{G} \operatorname*{max}_{C} \mathcal{V} (G, C),$}
\end{equation}

where $\mathcal{V}$ represents the standard GAN objective, allowing for adversarial training to enhance the robustness and generative capabilities of our model.

\subsubsection{Inference}

During the inference stage, our approach generates multiple procedure plans, denoted by $\tilde{\pi}^k = \tilde{a}^k_{1:T}$, for a specified time horizon $T$, by utilizing solely the initial and final observations. 

To achieve this, we draw $K$ latent noise vectors, $\mathit{z}^k$, and pass them through the generator, $G$, conditioned on a single start-goal observation. More formally, the generator is expressed as
\begin{equation}
\scalebox{1.05}{$\tilde{\pi}^k = {G}(v_{start}, v_{goal}, {z^k}, M, Cxt), \ \mathit{z}^k \sim \mathcal{N}(0, 1),$}
\end{equation}
where $k = 1, \dotsc, K$, and $M$ and $Cxt$ are the memory features and context features, respectively.

To obtain a probability distribution over actions at each timestep, $t$, of the plan, we calculate action frequencies as follows:
\begin{equation}\scalebox{1.15}{$
    \bar{\Pi} = \bar{a}_{1:T} = \frac{1}{K} \sum_{k=1}^{K} [\tilde{a}^k_1, \ldots, \tilde{a}^k_{T} ].$}
\end{equation}
Here, $a^k_t$ are one-hot vectors, and each $\bar{a}_t$ results in a marginal distribution over actions at a specific timestep, $t$.

However, most standard benchmark metrics for procedure planning require a single action sequence output for evaluation. To compute the most probable action sequence induced by our action distribution, $\bar{\Pi}$, we use the Viterbi algorithm~\cite{viterbi}.

In the context of first-order Markov model-based dynamic systems, the Viterbi algorithm has traditionally been employed to determine the most probable path. For a given sequence of length $T$ and $N$ possible states of the system, the Viterbi algorithm relies on two essential inputs: (i) a transition matrix, denoted as $A \in \mathbb{R}^{N \times N}$, which captures the probability of transitioning from one state, $a_i$, to another, $a_j$, and (ii) an emission matrix, denoted as $B \in \mathbb{R}^{T \times N}$, which describes the probability of each state, $a_i$, given a set of observations.

In our work, we utilize the marginal state probabilities for each timestep, denoted as $\bar{\Pi}$, as the emission matrix, $B$, and estimate the transition matrix, $A$, directly from the ground truth plans. Specifically, the transition matrix, $A$, is computed based on the frequencies of action co-occurrence in the training data. To calculate the value of $A_{i,j}$, we tally the occurrences of action $a_i$ being followed by action $a_j$ in the ground truth plans. Subsequently, we normalize each row of $A$ to sum to $1$, by applying $L_1$-normalization followed by a softmax operation with a temperature parameter $\tau=1$. Our Viterbi post-processing step can be interpreted as biasing sample selections from $\{\tilde{\pi}^k\}_{k=1}^{{K}}$ towards plans that are more likely under a first-order model of action transitions.

\section{Experiment Setups}

\subsection{Evaluation Metrics}

Similar to previous studies \cite{procedure2020, bi2021procedure, sun2021plate}, we assess the effectiveness of our approach using three metrics that increase in stringency. The first is mean Intersection over Union (mIoU), which evaluates the overlap between the predicted and ground truth action sequences as sets, regardless of their order, to determine whether the correct set of steps required to complete the plan is captured. Second, mean Accuracy (mAcc) compares the predicted and ground truth action sequences element-wise, taking into account the order of the actions. Third, Success Rate (SR) is only successful if the predicted plan is an exact match with the ground truth. 

Additionally, we measure the stochastic nature of our model using several probabilistic metrics. We compute the Kullback–Leibler (KL) divergence between our predicted plan distributions and the ground truth, measure how well the ground truth modes are represented by our results (Mode Recall), and how often our plans correspond to the ground truth mode (Mode Precision). To generate these metrics, we draw 1500 samples from our generative model for each $\{start, goal\}$ observation, explicitly approximating a distribution. We also include more standard probabilistic prediction metrics such as Negative Log Likelihood (NLL) and cosine distance in our evaluation.

\begin{table*}[t]
\centering %
{
\begin{tabular}{l c c c c }
\toprule %
& & \multicolumn{3}{c}{$T=3$}  \\ %
\cmidrule(l){3-5} 

Models & \textit{Supervision} & SR $\uparrow$ & mAcc $\uparrow$ & mIoU $\uparrow$     \\ %
\midrule %
Random & - & $<$0.01 & 0.94 & 1.66 \\ %
Retrieval-Based & - & 8.05 & 23.30 & 32.06  \\ %
WLTDO \cite{ehsani2018let} & - & 1.87 & 21.64 & 31.70  \\ %
UAAA \cite{abu2019uncertainty} & - & 2.15 & 20.21 & 30.87  \\ %
UPN \cite{srinivas2018universal} & V & 2.89 & 24.39 & 31.56 \\ %
DDN \cite{procedure2020} & V & 12.18 & 31.29 & 47.48  \\ %
Ext-GAILw/o Aug. \cite{bi2021procedure} & V & 18.01 & 43.86 & 57.16  \\ %
Ext-GAIL \cite{bi2021procedure} & V & 21.27 & 49.46 & 61.70  \\ %
P3IV~\cite{zhao2022p3iv}  & L &\textbf{22.58} & {51.26} & {73.07} \\ %
Ours & C &{21.74} & \textbf{52.49} &\textbf{74.99} \\ %
\bottomrule %
\end{tabular}
}
\caption{We present an evaluation of procedure planning results for the prediction horizon $T=3$ on CrossTask. The column name \textit{Supervision} indicates the type of state supervision employed during training. Specifically, we use the notation V and L to represent visual and language state representations, respectively.
}
\label{tab:crosstask_rst1}
\end{table*}

\begin{table*}[t]
\centering %
{
\begin{tabular}{l l c c c c}
\toprule %

Horizons & Methods & Sup. &   SR$\uparrow$ & mAcc$\uparrow$ & mIoU$\uparrow$ \\ %
\midrule
\multirow{5}*{$T=3$}
 & Random & -& $<$0.01 & $<$0.01 & 2.47  \\ %
 & Retrieval &  - & 4.38 & 17.40 & 32.06  \\ %
  & DDN~\cite{procedure2020} & V  & 13.9 & 20.19 & 64.78 \\ %
  & Ext-GAIL~\cite{bi2021procedure} & V & - & - & -\\ %
  & P3IV~\cite{zhao2022p3iv} & L & \textbf{14.48} & {22.86} & {73.15} \\
    & Ours & C & { 14.06 } & \textbf{ 26.90 } & \textbf{ 75.97 } \\
\bottomrule %
\end{tabular}
}
\caption{Procedure planning results on COIN~\cite{COIN} for prediction horizon $T = 3$. The column Sup denotes the type of state supervision applied in training. 
}
\label{tab:coin_rst1}
\end{table*}

\subsection{Implementation Details}

Our proposed model leverages pre-extracted language and vision features, obtained from a backbone model trained on the HowTo100M dataset for joint text-video embedding. The backbone model is capable of embedding both vision and language inputs into $512$-dimensional features. In our approach, we employ a transformer decoder with eight heads, two layers, and hidden states of dimensionality $128$. As the pre-trained features are of dimension $512$, we employ a MLP with a shape of $[512 \rightarrow 256 \rightarrow 128]$, interspersed with ReLU, to embed our initial features.

For the memory module, we empirically set the number of memory entries, denoted as $n$, to $128$. Regarding the caption embedding, we also employ a MLP with a shape of $[512 \rightarrow 256 \rightarrow 128]$. As for the critic, denoted as $C$, we utilize another three-layer MLP with a shape of $[256 \rightarrow 64 \rightarrow 32]$, where ReLU activations are applied in all layers. The dimensionality of the noise vector, denoted as $\mathit{z}$, is empirically set to $32$. Our model is trained for $200$ epochs on a single V4500 GPU, with an initial learning rate set to $7 \times 10^{-4}$, and decayed by a factor of $0.65$ every $40$ epochs.

To evaluate our model, we report the performance of the best-performing model on the validation set, which is randomly collected using 20\% of the training data. The final results on the test set are reported based on this selected model.

\section{Results and Analysis}




Table~\ref{tab:crosstask_rst1} and Table~\ref{tab:coin_rst1} present the evaluation results of our approach on the CrossTask and COIN datasets, respectively. Our method demonstrates competitive performance compared to state-of-the-art techniques, which confirms our assumption that previous approaches' intermediate supervision serves as contextual information.

Moreover, a recent study~\cite{wang2023association} indicates a disparity between text and image modalities in text generation tasks. It suggests that generating new text using the text domain is preferable to using the visual domain, which supports our use of captions in our approach.

Lastly, we hypothesize that the slight decline in some metrics' performance could be attributed to the noise present in the obtained captions. Therefore, further exploration of this noise and its impact on our method's performance is warranted.

\section{Conclusion}

Our work focuses on tackling the procedure planning task in instructional videos, and we propose a novel weakly-supervised setting that foregoes the use of intermediate supervision, including visual observations, language instructions, and task names. We believe that previous works' intermediate supervisions can provide valuable context information, and therefore, we utilize captions of visual start and goal observations as a cost-effective alternative form of supervision. We implement the BLIP technique to generate captions and use them as supervision to train the context feature with contrastive learning loss. Subsequently, the context feature is inputted into the generator to facilitate plan generation. Our experiments on two datasets with varying scales demonstrate that our model can achieve comparable performance on multiple metrics, which strongly supports our hypothesis.

\section{Future Directions}

\subsection{Procedure Planning in Complex Situation}

The process of cooking a dish can vary among individuals, with different people following different steps in different orders, and some steps being performed simultaneously, which introduces uncertainty into the planning process. ~\citet{wang2023pdpp} proposes to address this uncertainty by employing a diffusion model, which accounts for randomness during both training and sampling.

Despite the progress made by~\citet{wang2023pdpp}, there are still areas that could be improved. For instance, instead of outputting a sequence of plans, a more effective representation could be a structured flow graph that captures the diversity of plans. Another promising direction is to generate executable code for the plans, incorporating "if-else" and "for loop" commands that can better depict the complexity of plans~\cite{suris2023vipergpt}. Additionally, existing literature typically has a limited horizon of videos with few steps (<6), while real-life plans often involve more elaborate sequences. Furthermore, in practical scenarios, a single step may involve multiple instances or actions, whereas current approaches only consider one instance or action at a time. 

Therefore, developing a better representation of output plans, such as flow graphs or code, and creating a large-scale dataset that includes more steps and more complex single steps, would be promising avenues for future research.

\subsection{Procedure-Aware Representation Learning}

The task of representation learning for instructional videos has garnered significant attention in recent research. Zhou et al.\cite{zhou2023procedure} propose a method that builds a procedure knowledge graph from WikiHow and unlabeled videos, followed by pre-training the model with multiple objectives related to nodes and edges in the knowledge graph, thus enabling the representation to incorporate procedure knowledge. Zhong et al.\cite{zhong2023learning} introduce a diffusion process-based approach to model the temporal ordering of action steps, and pre-train the network using videos and their narrations. In contrast, Narasimhan et al.~\cite{narasimhan2023learning} adopt a classic masked reconstruction learning scheme, where the model is pre-trained with masked step modeling to make it aware of the task structure.

Although these methods have demonstrated superior performance, they still rely on existing databases such as WikiHow and the HowTo100M benchmark as the source of knowledge. In the era of large language models (LLMs), an exciting direction for future research is to explore how to efficiently distill procedure knowledge from LLMs, especially in terms of retrieving knowledge in an efficient manner. This could lead to advancements in representation learning for instructional videos by leveraging the capabilities of state-of-the-art LLMs.

\subsection{Video Demonstration}
VisualHow~\cite{yang2022visualhow} is proposed to complement WikiHow by introducing visual demonstrations for textual instructions. However, it is limited to the image domain, and a current trending topic is text-to-video generation. While current methods can generate visually-consistent components, such as PVDM~\cite{yu2023video}, they are unable to handle long-term video generation for demonstrations.

Instructional videos provide a natural scenario for this task, as they contain high semantic correspondence across the video. Short clips containing similar visual components can be generated based on the text description of each step using current methods. However, the challenge lies in connecting each clip to form a long video while considering the semantic relationships between them. Addressing this challenge is a valuable problem to be explored.

\subsection{Task Planning in Robotics using LLMs}

Another avenue of research that bears significant relevance to procedural learning is task learning in robotics. In recent years, language has emerged as a promising medium for addressing long-horizon robotics problems. A plethora of recent works has harnessed the generative capabilities of Large Language Models (LLMs) by employing them to predict long-horizon plans. For instance, Zeroshot-LLMs~\cite{zeroshot-llms-2022} ground an LLM planner to admissible action sets, albeit with evaluation limited to task-level planning. Other works have explored shifting the planning medium from natural language to code~\cite{code-as-policies-2022, progprompt-2022, zelikman2022parsel}, embedding task queries, robot actions, solution samples, and fallback behaviors as programs in the prompt.

Two seminal works in this domain are SayCan~\cite{saycan-2022} and Inner Monologue (IM)~\cite{innermono-2022}, which dynamically score the "usefulness" and "feasibility" of all possible skills at each timestep and execute the one with the highest score. Termination occurs when the score of the "stop" surpasses that of any other skill. IM provides additional sources of feedback to the LLM, such as object descriptions, skill successes, and task-progress cues.

Despite the demonstrated generality of SayCan and IM over a diverse range of tasks, there are several inherent limitations that hinder their performance. Firstly, by solely considering the next skill in a "greedy" manner at each timestep, they may fail to account for geometric dependencies that span across an entire skill sequence. Secondly, these methods do not account for the uncertainty of skill feasibility predictions, which has been established to be crucial in using skills to tackle geometrically complex task planning problems, as demonstrated in \cite{taps-2022}.

\bibliography{acl}

\begin{thebibliography}{36}
\expandafter\ifx\csname natexlab\endcsname\relax\def\natexlab#1{#1}\fi

\bibitem[{Abu~Farha and Gall(2019)}]{abu2019uncertainty}
Yazan Abu~Farha and Juergen Gall. 2019.
\newblock Uncertainty-aware anticipation of activities.

\bibitem[{Agia et~al.(2022)Agia, Migimatsu, Wu, and Bohg}]{taps-2022}
Christopher Agia, Toki Migimatsu, Jiajun Wu, and Jeannette Bohg. 2022.
\newblock Stap: Sequencing task-agnostic policies.
\newblock \emph{arXiv preprint arXiv:2210.12250}.

\bibitem[{Ahn et~al.(2022)Ahn, Brohan, Brown, Chebotar, Cortes, David, Finn,
  Gopalakrishnan, Hausman, Herzog et~al.}]{saycan-2022}
Michael Ahn, Anthony Brohan, Noah Brown, Yevgen Chebotar, Omar Cortes, Byron
  David, Chelsea Finn, Keerthana Gopalakrishnan, Karol Hausman, Alex Herzog,
  et~al. 2022.
\newblock Do as i can, not as i say: Grounding language in robotic affordances.
\newblock \emph{arXiv preprint arXiv:2204.01691}.

\bibitem[{Bi et~al.(2021)Bi, Luo, and Xu}]{bi2021procedure}
Jing Bi, Jiebo Luo, and Chenliang Xu. 2021.
\newblock Procedure planning in instructional videos via contextual modeling
  and model-based policy learning.

\bibitem[{Brown et~al.(2020)Brown, Mann, Ryder, Subbiah, Kaplan, Dhariwal,
  Neelakantan, Shyam, Sastry, Askell et~al.}]{GPT3}
Tom Brown, Benjamin Mann, Nick Ryder, Melanie Subbiah, Jared~D Kaplan, Prafulla
  Dhariwal, Arvind Neelakantan, Pranav Shyam, Girish Sastry, Amanda Askell,
  et~al. 2020.
\newblock Language models are few-shot learners.
\newblock 33:1877--1901.

\bibitem[{Chang et~al.(2020)Chang, Huang, Xu, Adeli, Fei-Fei, and
  Niebles}]{procedure2020}
Chien-Yi Chang, De-An Huang, Danfei Xu, Ehsan Adeli, Li~Fei-Fei, and
  Juan~Carlos Niebles. 2020.
\newblock Procedure planning in instructional videos.

\bibitem[{Ehsani et~al.(2018)Ehsani, Bagherinezhad, Redmon, Mottaghi, and
  Farhadi}]{ehsani2018let}
Kiana Ehsani, Hessam Bagherinezhad, Joseph Redmon, Roozbeh Mottaghi, and Ali
  Farhadi. 2018.
\newblock Who let the dogs out? {M}odeling dog behavior from visual data.

\bibitem[{Florensa et~al.(2018)Florensa, Held, Geng, and
  Abbeel}]{florensa2018automatic}
Carlos Florensa, David Held, Xinyang Geng, and Pieter Abbeel. 2018.
\newblock Automatic goal generation for reinforcement learning agents.

\bibitem[{Ghosh et~al.(2018)Ghosh, Gupta, and Levine}]{ghosh2018learning}
Dibya Ghosh, Abhishek Gupta, and Sergey Levine. 2018.
\newblock Learning actionable representations with goal-conditioned policies.
\newblock \emph{arXiv preprint arXiv:1811.07819}.

\bibitem[{Huang et~al.(2022{\natexlab{a}})Huang, Abbeel, Pathak, and
  Mordatch}]{zeroshot-llms-2022}
Wenlong Huang, Pieter Abbeel, Deepak Pathak, and Igor Mordatch.
  2022{\natexlab{a}}.
\newblock Language models as zero-shot planners: Extracting actionable
  knowledge for embodied agents.
\newblock \emph{arXiv preprint arXiv:2201.07207}.

\bibitem[{Huang et~al.(2022{\natexlab{b}})Huang, Xia, Xiao, Chan, Liang,
  Florence, Zeng, Tompson, Mordatch, Chebotar et~al.}]{innermono-2022}
Wenlong Huang, Fei Xia, Ted Xiao, Harris Chan, Jacky Liang, Pete Florence, Andy
  Zeng, Jonathan Tompson, Igor Mordatch, Yevgen Chebotar, et~al.
  2022{\natexlab{b}}.
\newblock Inner monologue: Embodied reasoning through planning with language
  models.
\newblock \emph{arXiv preprint arXiv:2207.05608}.

\bibitem[{Kaelbling(1993)}]{kaelbling1993hierarchical}
Leslie~Pack Kaelbling. 1993.
\newblock Hierarchical learning in stochastic domains: Preliminary results.

\bibitem[{Li et~al.(2022)Li, Li, Xiong, and Hoi}]{li2022blip}
Junnan Li, Dongxu Li, Caiming Xiong, and Steven Hoi. 2022.
\newblock Blip: Bootstrapping language-image pre-training for unified
  vision-language understanding and generation.
\newblock In \emph{International Conference on Machine Learning}, pages
  12888--12900. PMLR.

\bibitem[{Liang et~al.(2022)Liang, Huang, Xia, Xu, Hausman, Ichter, Florence,
  and Zeng}]{code-as-policies-2022}
Jacky Liang, Wenlong Huang, Fei Xia, Peng Xu, Karol Hausman, Brian Ichter, Pete
  Florence, and Andy Zeng. 2022.
\newblock Code as policies: Language model programs for embodied control.
\newblock \emph{arXiv preprint arXiv:2209.07753}.

\bibitem[{Mangalam et~al.(2020)Mangalam, Girase, Agarwal, Lee, Adeli, Malik,
  and Gaidon}]{mangalam2020not}
Karttikeya Mangalam, Harshayu Girase, Shreyas Agarwal, Kuan-Hui Lee, Ehsan
  Adeli, Jitendra Malik, and Adrien Gaidon. 2020.
\newblock It is not the journey but the destination: Endpoint conditioned
  trajectory prediction.

\bibitem[{Narasimhan et~al.(2023)Narasimhan, Yu, Bell, Zhang, and
  Darrell}]{narasimhan2023learning}
Medhini Narasimhan, Licheng Yu, Sean Bell, Ning Zhang, and Trevor Darrell.
  2023.
\newblock Learning and verification of task structure in instructional videos.
\newblock \emph{arXiv preprint arXiv:2303.13519}.

\bibitem[{Radford et~al.(2021)Radford, Kim, Hallacy, Ramesh, Goh, Agarwal,
  Sastry, Askell, Mishkin, Clark et~al.}]{clip}
Alec Radford, Jong~Wook Kim, Chris Hallacy, Aditya Ramesh, Gabriel Goh,
  Sandhini Agarwal, Girish Sastry, Amanda Askell, Pamela Mishkin, Jack Clark,
  et~al. 2021.
\newblock Learning transferable visual models from natural language
  supervision.
\newblock pages 8748--8763. PMLR.

\bibitem[{Radford et~al.(2019)Radford, Wu, Child, Luan, Amodei, and
  Sutskever}]{GPT}
Alec Radford, Jeff Wu, Rewon Child, David Luan, Dario Amodei, and Ilya
  Sutskever. 2019.
\newblock Language models are unsupervised multitask learners.

\bibitem[{Shi et~al.(2023)Shi, Darrell, and Wang}]{shi2023top}
Baifeng Shi, Trevor Darrell, and Xin Wang. 2023.
\newblock Top-down visual attention from analysis by synthesis.
\newblock \emph{arXiv preprint arXiv:2303.13043}.

\bibitem[{Singh et~al.(2022)Singh, Blukis, Mousavian, Goyal, Xu, Tremblay, Fox,
  Thomason, and Garg}]{progprompt-2022}
Ishika Singh, Valts Blukis, Arsalan Mousavian, Ankit Goyal, Danfei Xu, Jonathan
  Tremblay, Dieter Fox, Jesse Thomason, and Animesh Garg. 2022.
\newblock Progprompt: Generating situated robot task plans using large language
  models.
\newblock \emph{arXiv preprint arXiv:2209.11302}.

\bibitem[{Srinivas et~al.(2018)Srinivas, Jabri, Abbeel, Levine, and
  Finn}]{srinivas2018universal}
Aravind Srinivas, Allan Jabri, Pieter Abbeel, Sergey Levine, and Chelsea Finn.
  2018.
\newblock Universal planning networks: Learning generalizable representations
  for visuomotor control.

\bibitem[{Sun et~al.(2021)Sun, Huang, Lu, Liu, Zhou, and Garg}]{sun2021plate}
Jiankai Sun, De-An Huang, Bo~Lu, Yun-Hui Liu, Bolei Zhou, and Animesh Garg.
  2021.
\newblock {PlaTe}: Visually-grounded planning with transformers in procedural
  tasks.
\newblock \emph{arXiv preprint arXiv:2109.04869v1}.

\bibitem[{Sur{\'\i}s et~al.(2023)Sur{\'\i}s, Menon, and
  Vondrick}]{suris2023vipergpt}
D{\'\i}dac Sur{\'\i}s, Sachit Menon, and Carl Vondrick. 2023.
\newblock Vipergpt: Visual inference via python execution for reasoning.
\newblock \emph{arXiv preprint arXiv:2303.08128}.

\bibitem[{Tang et~al.(2019)Tang, Ding, Rao, Zheng, Zhang, Zhao, Lu, and
  Zhou}]{COIN}
Yansong Tang, Dajun Ding, Yongming Rao, Yu~Zheng, Danyang Zhang, Lili Zhao,
  Jiwen Lu, and Jie Zhou. 2019.
\newblock {COIN}: A large-scale dataset for comprehensive instructional video
  analysis.

\bibitem[{Tewel et~al.(2022)Tewel, Shalev, Schwartz, and
  Wolf}]{tewel2022zerocap}
Yoad Tewel, Yoav Shalev, Idan Schwartz, and Lior Wolf. 2022.
\newblock Zerocap: Zero-shot image-to-text generation for visual-semantic
  arithmetic.
\newblock pages 17918--17928.

\bibitem[{Viterbi(1967)}]{viterbi}
Andrew Viterbi. 1967.
\newblock Error bounds for convolutional codes and an asymptotically optimum
  decoding algorithm.
\newblock \emph{IEEE Transactions on Information Theory}, 13(2):260--269.

\bibitem[{Wang et~al.(2023{\natexlab{a}})Wang, Wu, Guo, and
  Wang}]{wang2023pdpp}
Hanlin Wang, Yilu Wu, Sheng Guo, and Limin Wang. 2023{\natexlab{a}}.
\newblock Pdpp: Projected diffusion for procedure planning in instructional
  videos.
\newblock \emph{arXiv preprint arXiv:2303.14676}.

\bibitem[{Wang et~al.(2023{\natexlab{b}})Wang, Yan, and
  Zhang}]{wang2023association}
Junyang Wang, Ming Yan, and Yi~Zhang. 2023{\natexlab{b}}.
\newblock From association to generation: Text-only captioning by unsupervised
  cross-modal mapping.
\newblock \emph{arXiv preprint arXiv:2304.13273}.

\bibitem[{Wu et~al.(2022)Wu, Luo, Fang, Wang, and Ouyang}]{wu2022cap4video}
Wenhao Wu, Haipeng Luo, Bo~Fang, Jingdong Wang, and Wanli Ouyang. 2022.
\newblock Cap4video: What can auxiliary captions do for text-video retrieval?
\newblock \emph{arXiv preprint arXiv:2301.00184}.

\bibitem[{Yang et~al.(2022)Yang, Chen, Jiang, Chen, Wang, and
  Zhao}]{yang2022visualhow}
Jinhui Yang, Xianyu Chen, Ming Jiang, Shi Chen, Louis Wang, and Qi~Zhao. 2022.
\newblock Visualhow: Multimodal problem solving.
\newblock In \emph{Proceedings of the IEEE/CVF Conference on Computer Vision
  and Pattern Recognition}, pages 15627--15637.

\bibitem[{Yu et~al.(2023)Yu, Sohn, Kim, and Shin}]{yu2023video}
Sihyun Yu, Kihyuk Sohn, Subin Kim, and Jinwoo Shin. 2023.
\newblock Video probabilistic diffusion models in projected latent space.
\newblock \emph{arXiv preprint arXiv:2302.07685}.

\bibitem[{Zelikman et~al.(2022)Zelikman, Huang, Poesia, Goodman, and
  Haber}]{zelikman2022parsel}
Eric Zelikman, Qian Huang, Gabriel Poesia, Noah~D Goodman, and Nick Haber.
  2022.
\newblock Parsel: A unified natural language framework for algorithmic
  reasoning.
\newblock \emph{arXiv preprint arXiv:2212.10561}.

\bibitem[{Zhao et~al.(2022)Zhao, Hadji, Dvornik, Derpanis, Wildes, and
  Jepson}]{zhao2022p3iv}
He~Zhao, Isma Hadji, Nikita Dvornik, Konstantinos~G Derpanis, Richard~P Wildes,
  and Allan~D Jepson. 2022.
\newblock P3iv: Probabilistic procedure planning from instructional videos with
  weak supervision.
\newblock In \emph{Proceedings of the IEEE/CVF Conference on Computer Vision
  and Pattern Recognition}, pages 2938--2948.

\bibitem[{Zhong et~al.(2023)Zhong, Yu, Bai, Li, Yan, and
  Li}]{zhong2023learning}
Yiwu Zhong, Licheng Yu, Yang Bai, Shangwen Li, Xueting Yan, and Yin Li. 2023.
\newblock Learning procedure-aware video representation from instructional
  videos and their narrations.
\newblock \emph{arXiv preprint arXiv:2303.17839}.

\bibitem[{Zhou et~al.(2023)Zhou, Mart{\'\i}n-Mart{\'\i}n, Kapadia, Savarese,
  and Niebles}]{zhou2023procedure}
Honglu Zhou, Roberto Mart{\'\i}n-Mart{\'\i}n, Mubbasir Kapadia, Silvio
  Savarese, and Juan~Carlos Niebles. 2023.
\newblock Procedure-aware pretraining for instructional video understanding.
\newblock \emph{arXiv preprint arXiv:2303.18230}.

\bibitem[{Zhukov et~al.(2019)Zhukov, Alayrac, Cinbis, Fouhey, Laptev, and
  Sivic}]{CrossTask}
Dimitri Zhukov, Jean-Baptiste Alayrac, Ramazan~Gokberk Cinbis, David Fouhey,
  Ivan Laptev, and Josef Sivic. 2019.
\newblock Cross-task weakly supervised learning from instructional videos.

\end{thebibliography}
\bibliographystyle{acl_natbib}

\end{document}